# SINGULAB – A GRAPHICAL USER INTERFACE FOR THE SINGULARITY ANALYSIS OF PARALLEL ROBOTS BASED ON GRASSMANN-CAYLEY ALGEBRA

Patricia Ben-Horin, Moshe Shoham

*Department of Mechanical Engineering*

*Technion – Israel Institute of Technology, Haifa, Israel*

[patbh,shoham]@tx.technion.ac.il

Stéphane Caro, Damien Chablat, Philippe Wenger

*Institut de Recherche en Communications et Cybernétique de Nantes*

*1, Rue de la Noe, 44321 Nantes, France*

[stephane.caro, damien.chablat, philippe.wenger]@irccyn.ec-nantes.fr

**Abstract**   This paper presents SinguLab, a graphical user interface for the singularity analysis of parallel robots. The algorithm is based on Grassmann-Cayley algebra. The proposed tool is interactive and introduces the designer to the singularity analysis performed by this method, showing all the stages along the procedure and eventually showing the solution algebraically and graphically, allowing as well the singularity verification of different robot poses.

## 1. Introduction

Singularity of parallel manipulators has been thoroughly investigated using different methods, mainly including line geometry, screw theory, and Jacobian determinants analysis. Recently, Grassmann-Cayley algebra (GCA) has been used for singularity analysis too.

SinguLab is the first version of a tool for singularity analysis of parallel robots. The aim of this user interface is to provide the designer an automatic tool for the analysis, geometric interpretation and visualization of singularities. It enables the user to determine the singularities of a large range of parallel robots and gives him some guidelines of GCA.

SinguLab was developed within the framework of SIROPA[1] – a French national project, the aim of which is to develop knowledge about the direct-kinematics singularities of parallel robots and to transmit this knowledge to the end-users – during a sojourn stay of the first author at IRCCyN.

## 1.1 Grassmann-Cayley algebra

The algorithm used in SinguLab is based on GCA. For space limitations we only introduce the basic concepts and the readers are referred to (Ben-Horin and Shoham, 2006a) and reference therein for further details on this topic. The basic elements of this algebra are called *extensors*, which in fact are symbolically denoted Plücker coordinates of vectors. Two basic operations that play an essential role in GCA involving extensors are the *join* and *meet* operators. The first is associated with the union of two vector spaces, and the latter has the same geometric meaning as the intersection of two vector spaces. Further, special determinants called *brackets* are also defined in GCA. The brackets, of which columns are vectors, satisfy special product relations called *syzygies*, which are useful to manipulate and compare bracket expressions. The Grassmann-Cayley algebra functions under the projective space $P^d$, in which points are represented by homogeneous coordinates and lines are represented by Plücker coordinates.

As mentioned above, the extensors are vectors that represent geometric entities, and are characterized by their *step*. Extensors of step 1, 2 and 3 stand for a point, a line and a plane, respectively. Assuming two extensors **A** and **B**, of step $k$ and $h$, respectively, defined in the $d$-projective space, the *join* and *meet* operations are written as follows:

$$\mathbf{A} \vee \mathbf{B} = a_1 \vee a_2 \vee \cdots \vee a_k \vee b_1 \vee \cdots \vee b_h = a_1 a_2 \cdots a_k b_1 \cdots b_h \quad (1)$$

$$\mathbf{A} \wedge \mathbf{B} = \sum_\sigma \text{sgn}(\sigma)[a_{\sigma(1)} a_{\sigma(2)} \ldots a_{\sigma(d-h)} b_1 \ldots b_h] a_{\sigma(d-h+1)} \ldots a_{\sigma(k)}$$
$$= \sum_\sigma \text{sgn}(\sigma)[\dot{a}_1 \dot{a}_2 \ldots \dot{a}_{d-h} b_1 \ldots b_h] \dot{a}_{d-h+1} \ldots \dot{a}_k \quad (2)$$

where the sum in Eq.(2) is taken over all permutations σ of {1,2,..,$k$} such that σ(1)<σ(2)<…<σ($d$-$h$) and σ($d$-$h$+1)<σ($d$-$h$+2)<…<σ($k$). Incidences between geometric entities are obtained as extensors of step 0 (scalars). Some examples of incidences in 3D-space are the *meet* of four planes, the *meet* of two lines and the meet of a line with two planes. Three *meet* examples are written in GCA as follows:

---

[1] https://wiki-sop.inria.fr/wiki/bin/view/Coprin/SIROPA

Meet of four planes: $abc \wedge def \wedge ghi \wedge jkl = \left[\dot{a}\,def\right]\left[\dot{b}\,ghi\right]\left[\dot{c}\,jkl\right]$

Meet of two lines: $ab \wedge cd = [abcd]$

Meet of a line and two planes: $gh \wedge abc \wedge def = \left[\dot{g}\,abc\right]\left[\dot{h}\,def\right]$

Let us consider a finite set of 1-extensors $\{a_1, a_2, ., a_d\}$ defined in the $d$-dimensional vector space over the field $\Upsilon$, $V$, where $a_i = x_{1,i}, x_{2,i}, .., x_{d,i}$ ($1 \leq i \leq d$). The *bracket* of these extensors is the determinant of the matrix, of which columns are vectors $a_i$ ($1 \leq i \leq d$):

$$[a_1, a_2, ..., a_d] = \begin{vmatrix} x_{1,1} & x_{1,2} & \cdots & x_{1,d} \\ \vdots & \vdots & \cdots & \vdots \\ x_{d,1} & x_{d,2} & \cdots & x_{d,d} \end{vmatrix}, \qquad (3)$$

For example, the bracket of points $a$, $b$, $c$ and $d$ defined in the 3D space is written as:

$$[\mathbf{abcd}] = \begin{vmatrix} a_x & b_x & c_x & d_x \\ a_y & b_y & c_y & d_y \\ a_z & b_z & c_z & d_z \\ 1 & 1 & 1 & 1 \end{vmatrix}$$

From a geometrical point of view, the value of this bracket represents six times the volume of the tetrahedron of vertices **a**, **b**, **c** and **d**.

## 2. Algorithm

The procedure behind SinguLab follows the next steps:
a. Determination of the robot structure. *b.* Writing the singularity equation in terms of GCA. *c.* Identification of the geometrical entities involved in the singularity condition according to the algebraic equation. *d.* Depending on the entities found, the algorithm finds the geometrical condition in terms of GCA. The singularity condition, if feasible, is shown to the user by means of a geometrical statement in algebraic form with a graphical visualization of the geometric entities comprising the singularity.

### 2.1 Determination of the robot structure

The available options in this version are all the possible Gough-Stewart platforms (GSPs). There are 35 different GSPs if concurrent joints on the platform or on the base are considered (Faugere and Lazard, 1995). In the next version of SinguLab, other types of parallel robots will be analyzed by means of the method explained in sections 2.5 and 3.

The robot structure is determined by the user with six lines as 2-extensors according to their endpoints on the platform and the base. Two concurrent joints have the same label. Once the structure is defined, a schematic of the robot appears.

## 2.2 Singularity equation

The singularity analysis is performed using a coordinate-free invariant version of the Jacobian matrix determinant written in terms of GCA, which is suitable for robots of motion ruled by six pure forces, represented by six zero-pitch screws. This coordinate-free version of the Jacobian determinant was derived by McMillan and White (1991), after proposing a significantly larger expression by White (1983) eight years before.

A paradigm of robots ruled by six pure forces is the general GSP. The moving platform of GSP is connected to the base through six spherical-prismatic-universal chains, the spherical and universal joints being placed anywhere on both platforms. The coordinate-free version of the Jacobian determinant of this robot with legs *ab*, *cd*, *ef*, *gh*, *ij*, and *kl* (*a,..,l* denoting the endpoints of the lines) has 24 monomials as follows:

$$\begin{aligned}[[ab,cd,ef,gh,ij,kl]] = &-[abcd][efgi][hjkl]+[abcd][efhi][gjkl]+[abcd][efgj][hikl]\\&-[abcd][efhj][gikl]+[abce][dfgh][ijkl]-[abde][cfgh][ijkl]-[abcf][degh][ijkl]\\&+[abdf][cegh][ijkl]-[abce][dghi][fjkl]+[abde][cghi][fjkl]+[abcf][dghi][ejkl]\\&+[abce][dghj][fikl]-[abdf][cghi][ejkl]-[abde][cghj][fikl]-[abcf][dghj][eikl]\\&+[abdf][cghj][eikl]+[abcg][defi][hjkl]-[abdg][cefi][hjkl]-[abch][defi][gjkl]\\&-[abcg][defj][hikl]+[abdh][cefi][gjkl]+[abdg][cefj][hikl]+[abch][defj][gikl]\\&-[abdh][cefj][gikl]\end{aligned}$$

(4)

Each term (monomial) in Eq.(4) is a multiplication of three brackets.

The singularity condition arises when the right-hand side of (4) is equal to zero. This is the basic equation used in the singularity analysis in this paper. It is to be noted that we use Eq.(4) instead of another shorter version having 16 monomials (Downing et al, 2002). The main advantage of Eq. (4) over the shorter version is the order of the points in each bracket, which is lexicographically in both rows and columns. This fact significantly facilitates the manipulation and comparison of monomials, operations needed for the derivation of the geometric condition of the singularity equation.

For the remaining 34 GSP combinations, the singularity equation is significantly reduced since many monomials vanish due to the appearance of equal points in some brackets. For most of these structures, this equation enables the geometrical explanation of the singularity condition using GCA tools. The reduction of the original equation, however, may be differently obtained if different order of legs is taken to substitute the left-hand-side of Eq.(4). For example, for the 3-3 GSP, the following leg definition leads to two and four monomials if the leg order is altered:

$$[[ab,af,cb,cd,ed,ef]]=[abfc][acde][bdef]+[abfd][acbe][cdef] \qquad (5)$$

$$[[ab,cd,af,cb,ed,ef]]=-[abcd][afce][bdef]+[abcd][afbe][cdef]$$
$$+[abcf][dcbe][adef]-[abdc][cafe][bdef] \qquad (6)$$

Although Eqs.(5) and (6) are equivalent, the lowest number of monomials is recommended in order to avoid long calculations. Therefore, if the number of monomials obtained after the user definition is more than 4, then he is led to use the automatic function to find the shortest form. This function runs all the possible orders and returns the first shortest form. The next step is to find the interchangeable points within the monomials in order to identify the geometrical entities involved in the singularity condition.

### 2.3  Identification of interchangeable points

The objective of this stage is to automatically find the geometric entities involved in the singularity condition. These entities may be lines, planes or tetrahedrons. The method to find them is based on the first stage of the Cayley factorization performed by White (1991).

White's algorithm deals only with multilinear expressions, which are those containing each point in each monomial only once (Eq.(4) without any substitution is a multilinear example). According to his algorithm a pair of points is interchangeable if the expression after replacing all the appearances of both with each other, summed with the original expression is equal to zero:

$$P(a,b,...)+P(b,a,....)=0 \qquad (7)$$

where $P$ is the expression containing all the monomials (for example, Eq.(4)). This process is performed for all possible pairs of points, using the straightening algorithm (White, 1991).

Unfortunately, our expressions are never multilinear. Unlike the general GSP, of which singularity has no special geometrical explanation with this method, all other structures have at least one point appearing at least twice in each monomial. Until now, no algorithm of Cayley factorization for non-multilinear polynomials is known and it still remains an open problem. Our approach is as follows.

First, we assume that if a point appears more than once in each monomial, then each appearance belongs to a different geometric entity. Each monomial has three brackets, each bracket containing four points, thus twelve points are part of geometric entities that have to be identified. From the definition of the *meet* operation (Eq.(2)), to obtain a monomial of brackets of four points the geometric entities involved may be 2- or 3-extensors (lines or planes). Otherwise, a *meet* including a 4-extensor (tetrahedron) and another entity would lead to a 5-bracket. Still, a monomial of 4-brackets may result from a bracket containing a tetrahedron and two other brackets resulting from a meet of lines and planes. Accordingly, when the potential entities to be searched are lines, planes and tetrahedrons, the following groups can be found: *a*. Six lines; *b*. two planes and three lines; *c*. four planes; *d*. one tetrahedron, two planes and one line; *e*. one tetrahedron and four lines; *f*. two tetrahedrons and two lines; *g*. three tetrahedrons. To avoid the non-multilinearity problem, the order of searching is as follows:

1. The first entities to be searched are the tetrahedrons. These are searched as common brackets (having the four points of the tetrahedron) in all the monomials. If the equation has more than one monomial, then it is searched if there is a common bracket appearing in all the monomials. If such a bracket exists, then the tetrahedron recognition is done, and the remaining equation continues the search procedure. To have three tetrahedrons there has to be only one monomial in the equation, where each bracket consists of the points of each tetrahedron. If this is the case, then the procedure is completed, resulting in three possible coplanar tetrahedrons, according to the points appearing in each bracket in the monomial.

2. The second stage is to look for planes, which are represented as triplets of interchangeable points. These triplets are searched as points that appear together in one bracket in every monomial. Any pair within such a triplet is interchangeable since replacing them one by another means a permutation, leading to a general sign change in all the monomials, and thus satisfying Eq.(7).

3. Once all the triplets in the previous stage were recognized, their labels acquire a star ($a \rightarrow a^*$) to distinguish them from the other same labels appearing in other brackets in the monomials, due to the non-multilinearity. The next stage consists in searching among the remaining points, pairs that satisfy Eq.(7). In many cases it is not necessary to use the straightening algorithm to verify this condition. For the cases in which it would be necessary, unfortunately we cannot use this algorithm because of the non-multilinearity of the expressions. Therefore in these cases the points that remain without being identified to any entity are left in parentheses and are treated as follows. If three planes were already identified then the residual letters will be treated as possible part of a fourth plane (case *c*). The same occurs with the residual letter if three planes and one line were found. If two planes and a line were identified, then the residual letters will be referred as possible pairs of lines to correspond to case *b*.

## 2.4    Singularity solution and visualization

This stage provides the singularity condition as a geometrical incidence between the entities that were already identified. The *union* and *intersection* of geometric entities in terms of GCA are obtained by means of the *join* and *meet* operators, respectively.

Ben-Horin and Shoham (2006a, 2006b, 2007) found the geometrical conditions for 31 from the 34 regular GSPs. According to the number and types of entities identified in the previous section, this stage verifies if, algebraically, the respective condition is equal to a geometric incidence, some examples of which are shown in section 1. If both are equal, then the solution is written according to the points that were defined before, and the geometric entities involved are shown in the robot figure.

Once the singularity condition is obtained, a new field appears in the window with a value standing for the singularity condition in the actual configuration. When this value, which changes depending on the robot pose, becomes smaller than a predefined ε, it means closeness to singularity and a warning message appears. In the robot figure the user is able to move the platform in 6 DOFs in order to visualize the poses that satisfy the singularity condition. The singularity test is performed using the simplest form according to the condition obtained. For example, the four intersecting planes condition is verified by calculating the condition number of the 4×4 matrix containing the coefficients of the planes. The condition for two planes to intersect simultaneously a line is verified by calculating the condition number of the matrix having the four homogeneous coordinates of four points: two of them lie on the line of intersection of the planes, and the remaining two lie on the other line.

For the interested users, the identification of the interchangeable points and the verification of the singularity condition can be performed manually. By default, the automatic mode is applied.

## 2.5    Applications

The structures suitable to be analyzed by this method and used with this tool include a long list of robots in addition to all the GSP structures. Their analyses are performed knowing the equivalent lines of action applied to the platform by the legs. For six degree-of-freedom (DOF) robots these equivalent lines of action are the reciprocal screws to the passive joints of legs (McCarthy, 2000). The topology of the lines of action must be equal to one of the 34 combinations of legs arrangement presented in (Faugere and Lazard, 1995) since they are all the different combinations existing to define six legs connecting two platforms when two and three concurrent joints are taken into consideration.

Manipulators with lower-DOF having a spherical joint in each leg can be also analyzed by this tool. For their analysis the 6×6 Jacobian matrix, which contains a Jacobian of actuations and a Jacobian of constraints as sub-matrices, is needed (Joshi and Tsai, 2002). Once the rows of both sub-matrices are identified, these rows actually being the wrench screws applied to the platform, the topologically equivalent GSP can be identified and the singularity condition can be found by means of SinguLab.

In a future version of SinguLab, an automatic specification of other parallel robots than GSPs with graphical tools will be available. The use of SinguLab is performed according to the interactive instructions in the main window. These instructions follow the steps listed in Section 2. This guidance, showing each result algebraically, graphically and as a geometric statement, enables non-experts to obtain the singularity in an easy manner.

## 3. Example

We exemplify the tool by means of the singularity analysis of a 4-DOF robot, first presented in (Gallardo-Alvarado J. et al, 2006). This robot consists of three different legs, having PS, UPS and PRPS (or CPS) kinematic chains, respectively, (the underlined labels stand for actuated joints) see Fig.1(a).

The constraint screws of leg 1 are reciprocal to the spherical and prismatic joints. Thus they form a two-system perpendicular to the prismatic and passing through the spherical joint. Then the reciprocal screw to the passive joint of leg 1, $\$_1$, is a screw directed along the leg. In leg 2 the passive joints are the spherical and the universal joints thus their reciprocal screw, $\$_2$, is directed along the leg that connects them. The third leg has two prismatic actuators $P_3$ and $P_4$. The reciprocal screws to the passive joints form a two-system of zero pitch, being a planar pencil with center at the spherical joint and containing the axis of the revolute joint. Particularly, $\$_3$ and $\$_4$ pass through the spherical joint center and are directed along the leg and parallel to the revolute joint axis, respectively. With these screws known, the equivalent structure to be entered into SinguLab is as appears in Fig. 1(c). The interface analyzing this robot is shown in Fig.2, where the singularity condition is that at least one of the tetrahedrons composed by $S_1A_1B_1C_1$, $S_1S_3A_3B_3$ and $S_1S_3S_2A_2$ (according to the labels in Fig.1(c)) is coplanar.

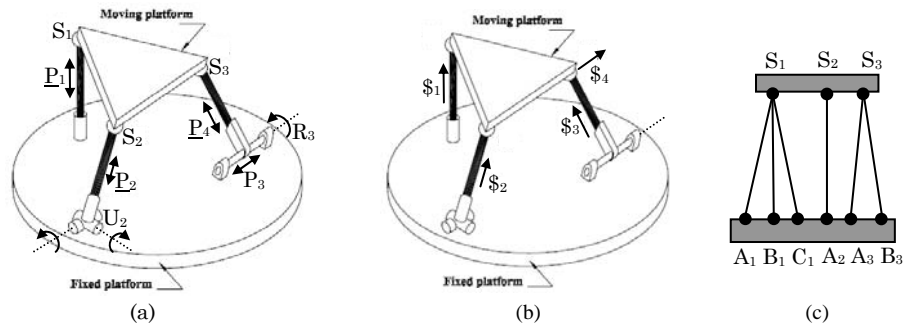

*Figure 1.* (a). Four-DOF robot from (Gallardo-Alvarado J. et al, 2006), (b). Reciprocal screws, (c). Equivalent structure.

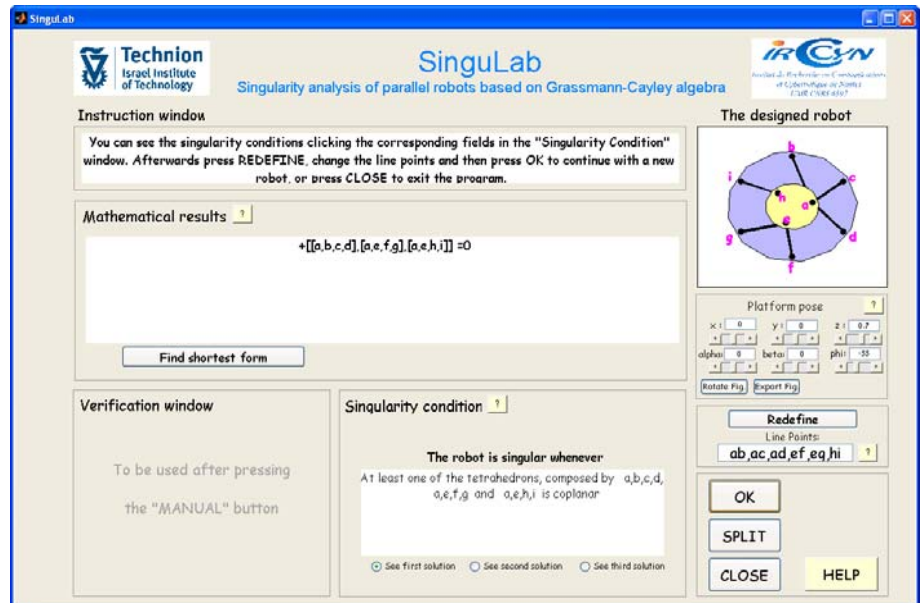

*Figure 2.* SinguLab interface

## 4. Conclusion

This paper presents SinguLab, a graphical user interface for the singularity analysis of parallel robots. The theoretical background behind this analysis is based on Grassmann-Cayley algebra, which provides a coordinate-free approach for treating geometric entities and their incidences. The identification of the geometric entities and the singularity conditions are made automatically. The results are based on previous studies on certain classes of Gough-Stewart platforms. This interface is suitable for a broad range of parallel robots. The current version can be used only for robots actuated with SPS chains. For those with serial chains, an equivalent structure has to be predefined with their reciprocal screws Accordingly, the topological arrangement of the lines of action is the robot definition input. For lower-DOF parallel robots, the reciprocal screws standing for the actuation and for the constraints on the platform have to be identified first. Therefore, we come up with the 6×6 Jacobian matrix and the same GCA approach thereafter. In future versions, the analysis of these robots will be incorporated in the software to provide a fully automatic tool for robot designers.

## 5. Acknowledgment

This work is supported in part by the French Research Agency A.N.R. (Agence Nationale pour la Recherche).